\documentclass[a4paper,conference]{IEEEtran}      

\usepackage{graphicx}
\usepackage[export]{adjustbox}
\usepackage{float}
\usepackage{multicol}
\usepackage{lipsum}
\usepackage{amssymb}
\usepackage{multirow}
\usepackage{amsmath}
\usepackage{tikz}
\usepackage{ctable}
\usepackage[font=small,labelfont=bf]{caption}
\usepackage{subcaption}
\usepackage{pgfplots}
\usepackage{filecontents}

\title{\LARGE \bf
Driver Intention Anticipation Based on In-Cabin and Driving Scene Monitoring} 

\author{
    Yao Rong\\
    Human-Computer Interaction\\
    University of T{\"u}bingen\\
    yao.rong@uni-tuebingen.de
    \and
    Zeynep Akata\\
    Explainable Machine Learning\\
    University of T{\"u}bingen\\
    zeynep.akata@uni-tuebingen.de
    
  \and
    Enkelejda Kasneci\\
    Human-Computer Interaction\\
    University of T{\"u}bingen\\
    enkelejda.kasneci@uni-tuebingen.de

}

\begin{document}

\maketitle
\thispagestyle{empty}
\pagestyle{empty}

\begin{abstract}
Numerous car accidents are caused by improper driving maneuvers. Serious injuries are however avoidable, if such driving maneuvers are detected beforehand and the driver is assisted accordingly. In fact, various recent research has focused on the automated prediction of driving maneuver based on hand-crafted features extracted mainly from in-cabin driver videos. Since the outside view from the traffic scene may also contain informative features for driving maneuver prediction, we present a framework for the detection of the drivers' intention based on both in-cabin and traffic scene videos. More specifically, we (1) propose a Convolutional-LSTM (ConvLSTM)-based auto-encoder to extract motion features from the out-cabin traffic, (2) train a classifier which considers motions from both in- and outside of the cabin jointly for maneuver intention anticipation, (3) experimentally prove that the in- and outside image features have complementary information. Our evaluation based on the publicly available dataset Brain4cars shows that our framework achieves a prediction with the accuracy of 83.98\% and $F_1$-score of 84.3\%. 
\end{abstract}

\section{INTRODUCTION}

\begin{figure*}[h]
 \centering
  \includegraphics[width=0.9\linewidth]{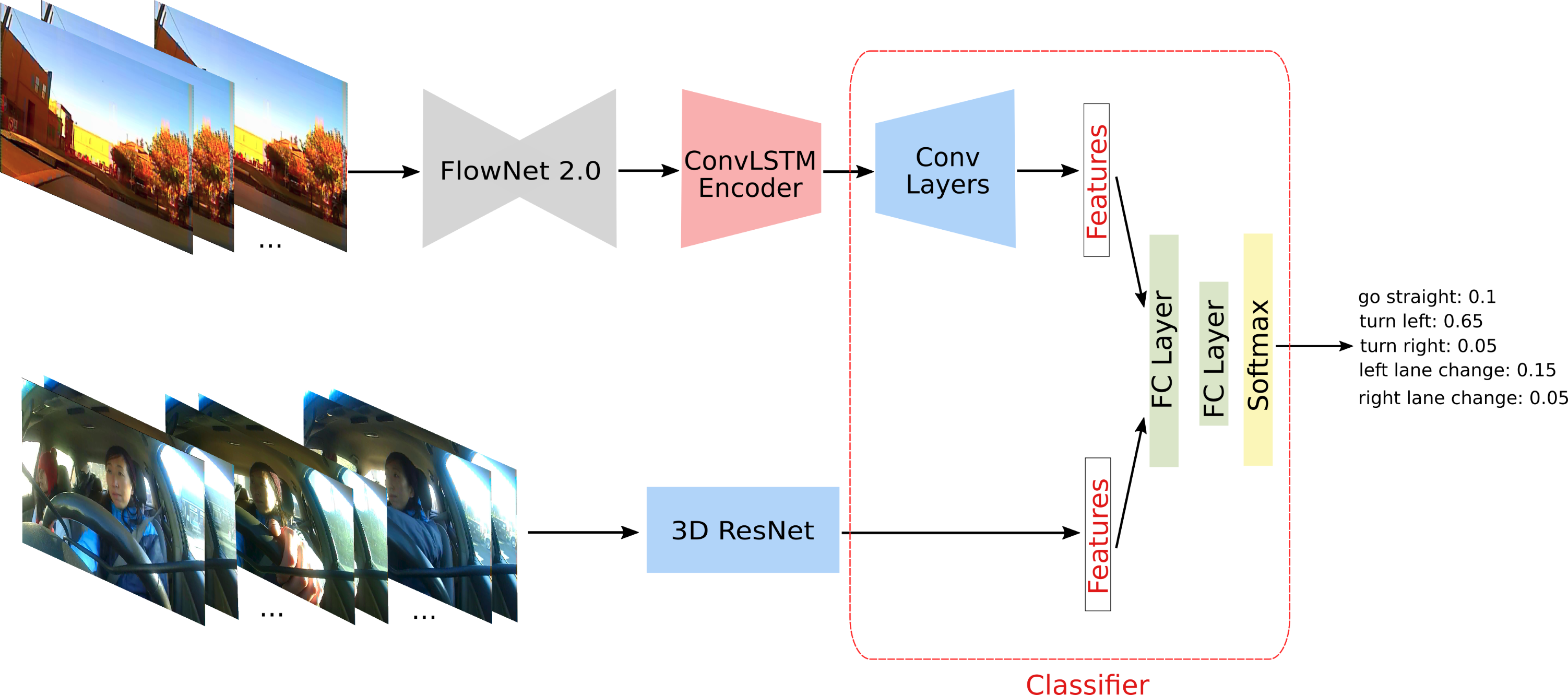}
  \caption{The overview of our framework. The upper branch depicts the feature extraction from out-cabin videos: FlowNet 2.0 extracts the optical flow from the consecutive frames; then the traffic motion is captured by a ConvLSTM-based encoder. The bottom branch represents the feature extraction from in-cabin videos based on the 3D ResNet-50 network. The red frame in the end refers to the classifier, where a decoder (marked as ``Conv Layers") for outside features is integrated. This novel classifier architecture allows features from in- and outside of the cabin to be considered jointly.}
  \label{fig: overview}
\end{figure*}

According to the World Health Organization \cite{WHO}, about 1.35 million people die in car accidents every year worldwide. These statistics, however, do not include non-fatal injuries from traffic accidents. Most of these accidents are caused by improper driver behavior: Based on the statistics from the Department for Transport (DfT) in Great Britain, a survey \cite{dftuk} revealed that there were 15,560 accidents reported due to poor turn or maneuver, which ranked top 5 in causes of road accidents in 2017. As automated vehicle technology emerges, it promised to be safer than human driving \cite{morando2018studying, fox2014self, teoh2017rage}. However, there is still much research to be conducted in order to reach to the fully automated level working at any possible traffic situation and weather conditions. On the half way to autonomous driving vehicles, it is therefore necessary to provide already existing Advanced Driver Assistance Systems (ADAS) the functionality for collaboration with the human driver in the most efficient way, for example to alert the driver in case of a dangerous maneuver.

Recently, many researchers focused on detecting maneuver intention of the driver before execution. For example, Brain4cars \cite{jain2015car} and  Honda Research Institute Driving Dataset (HDD) \cite{ramanishka2018toward} are two datasets specifically designed for learning driver behaviors. 
HDD for example \cite{ramanishka2018toward} uses three high-resolution video cameras, GPS, signals from LiDAR sensor and vehicle CAN-Bus to record the traffic scenes. Brain4cars \cite{jain2015car} provides videos from inside and outside of the car. GPS and vehicle dynamics are also recorded with the videos. These videos show different behavior patterns of maneuvers from driver side and road traffic. Images convey massive information, and much of the literature shows the possibility to predict driver intention according to the drivers' videos, since the drivers turn their heads to glance in the side mirrors. Previous work based on the Brain4cars dataset, such as \cite{jain2015car, jainbrain4cars, zhou2018driving, gebert2019end, tonutti2019robust}, have all achieved maneuver prediction. Although the reported results are quite impressive, there are still some issues that deserve scrutiny. 

More specifically, most of the previous works in the driver maneuver prediction domain mainly use videos from driver observation. Various research has shown that driver behavior, and especially eye movements of the driver, can not only be used for activity recognition \cite{braunagel2015driver,braunagel2017online} but also to ensure safe take-over behavior in conditionally autonomous driving \cite{braunagel2017ready}. Additionally, video frames of driver observations are used to extract features e.g. head postures \cite{jain2015car, jainbrain4cars, zhou2018driving, tonutti2019robust}. However, in these works, the traffic information is manually encoded into a vector with four elements, where the first two Boolean values indicate whether a lane exists on the right or left side of the vehicle, the third bit (also Boolean) implies if an intersection or turn exists in 15 meters, and the last value represents the current speed of the car. Therefore, video information of the outside view is not further processed. In addition, manual encoding as employed so far is not applicable to  practical use-cases. (2) \cite{gebert2019end} proposes using two 3D ResNet-101 models for two streams separately. However, it shows that using only driver videos works better than using both video streams. The reason behind this poor performance of outside videos is that there is no large dataset for on-road traffic training, which makes  training with the Brain4cars dataset from scratch very difficult. In contrast, for driver observation videos, there is large human activity dataset available such as Kinetics \cite{carreira2017quo}. 

Intuitively, the outside video, i.e., the scene perspective, should be very informative and  provide information that the inside video does not convey. Therefore, our work aims (1) extracting the vehicle motion information from the traffic videos effectively and improving the results which only used one video stream; (2) proposing an end-to-end method without using manual encoding information, and (3) keeping the model as light-weighted (less parameters) as possible to offer applicability to resource-limited mobile platforms.

To approach these aims, we propose a deep learning framework, which combines the information from the driver monitoring videos with the outside view. This framework is shown in Fig. \ref{fig: overview}. In our framework, a ConvLSTM \cite{xingjian2015convolutional} based encoder (shown in upper branch) extracts the motion information, which is interpreted in optical flow images. Meanwhile, the 3D ResNet-50 (shown in bottom branch) acquires features from the driver video. The motion decoder for outside motion features is integrated in the classifier. This novel classifier leverages features from both sides, i.e., driver and scene, jointly to produce a maneuver anticipation.

The contribution of our work is manifold: (1) we encode the traffic scene motion using a ConvLSTM-based auto-encoder, (2) propose a deep net framework investigating features from two incoming streams (in- and outsides) jointly, without using any manual-encoded or hand-crafted information, (3) achieve a state-of-the-art maneuver anticipation performance with less parameters compared to the previous work \cite{gebert2019end}, and (4) experimentally validate that the in- and outside videos contain complementary information.

The remaining of this paper is organized as follows: In Section \ref{sec:related work}, we first discuss related works. Our proposed methods and modules mentioned in Fig. \ref{fig: overview} are explained in detail in Section \ref{sec:methodology}. In Section \ref{sec:exp}, we introduce the dataset used for training and evaluation of our method and discuss our evaluation results. Finally, we summarize our main findings and conclude this paper.

\section{RELATED WORK}
\label{sec:related work}
Maneuver intention can be detected from drivers' behaviors, such as looking at the outside mirrors or out of the windows. Therefore, popular methods from the  domain of human action recognition are suitable and have been applied to tackle this challenge. An action consists of spatial and temporal information. As widely known, features in the spatial domain can be captured by Deep Convolutional Neural Networks (CNNs), while Recurrent Neural Network (RNN) architectures and Long Short-Term Memory (LSTM) cells are well-known for comprehending the logic hidden in time series. 
LSTM and RNN techniques are therefore often combined with 2D CNNs in video processing applications to deal with both spatial and temporal information, for example as in \cite{xingjian2015convolutional}.
 The formulation from \cite{xingjian2015convolutional} is shown in Eq. \ref{lstm eq} with a minor modification, since it contains no bias component. 
\begin{equation}
\label{lstm eq}
\begin{aligned}
\centering
&i_t = \sigma (W_{xi}*x_t + W_{hi}*h_{t-1} + W_{ci}\cdot
c_{t-1})\\
&f_t = \sigma (W_{xf}*x_t + W_{hf}*h_{t-1} + W_{cf}\cdot
c_{t-1})\\
&g_t = tanh(W_{xc}*x_t + W_{hc}*h_{t-1})\\
&c_t = f_t \cdot c_{t-1} + i_t \cdot g_t\\
&o_t = \sigma (W_{xo}*x_t + W_{ho}*h_{t-1} + W_{co}\cdot c_t)\\
&h_t = o_t \cdot \tanh(c_t)\\
\end{aligned}
\end{equation}
In the above Eq. \ref{lstm eq}, subscript $t$ implies the time sequence. $x_t$ is the input. $i_t, g_t, f_t$ and $o_t$ are the gates in the cell. $c_t$ is the cell state and $h_t$ is the hidden state. All the $W$s refer to the weights in a convolutional operation. $*$ denotes the convolution operation, while $\cdot$ refers to the element-wise multiplication. $\sigma$ and $\tanh$ are sigmoid and hyperbolic tangent functions, respectively, which are also applied element-wise.
The features learned by ConvLSTM can be used for regression or classification problems. For instance, the authors from \cite{xingjian2015convolutional} built an encoding-forecasting structure to predict the future frame using ConvLSTM cells.

One essential element of video analyzing is motion understanding. Motion describes changes in both temporal and spatial spaces and is often estimated on an image plane based on the optical flow. This technique has been researched for decades since \cite{horn1981determining}. It calculates the motion of individual pixels in consecutive frames, which can be then aggregated to interpret the motion of objects. Optical flow is for example widely used in automobile applications \cite{menze2015object}, since it serves as an extra feature. The extraction of optical flow has been regarded as an optimization problem in the past with various approaches for optical flow estimation such as energy-based method \cite{adelson1985spatiotemporal}, or region-based matching \cite{anandan1989computational}. However, with the rapid development of deep learning, CNN-based networks achieved very impressive results. \cite{dosovitskiy2015flownet, ilg2017flownet} are only two representative networks for this problem performing in an  end-to-end style, where the networks take two consecutive frames as input and output the optical flow.

As previously mentioned, there are multiple works aiming at the driver maneuver anticipation \cite{jain2015car, jainbrain4cars, zhou2018driving, gebert2019end, tonutti2019robust}. However, none of the previous work solved driver intention prediction with information from both video (in and out of the car) streams, since the traffic on road is too complex for hand-crafting explicit features. Therefore, several works, such as  \cite{jain2015car, jainbrain4cars, zhou2018driving, tonutti2019robust}, use manual-encoded feature vectors. On the other hand, training CNNs with outside videos in an end-to-end fashion did not show satisfactory results  \cite{gebert2019end}, since there was not enough on-road video data related to maneuver anticipation for training a CNN-based deep network. 

In contrast to the above mentioned approaches, we propose to use the outside video stream and the driver observation data jointly for intention anticipation. In the following sections, we introduce our method that leverages information from both videos towards an accurate intention anticipation.

\section{METHODOLOGY}
\label{sec:methodology}

\subsection{Future Frame Prediction}
\label{sec:frame prediction}

\begin{figure}[b!]
 \centering
  \includegraphics[width=\linewidth]{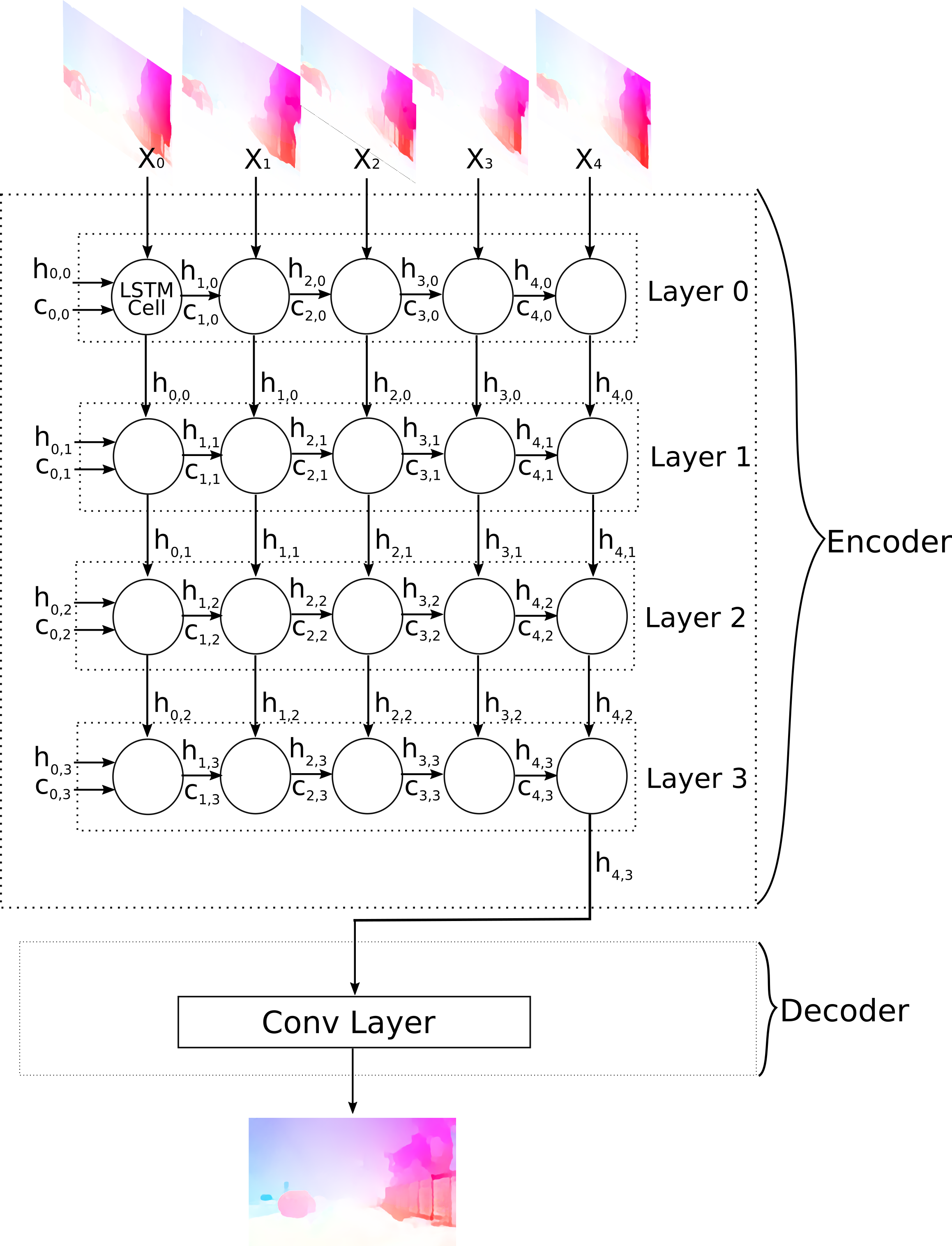}
  \caption{Architecture of the proposed future motion prediction module.}
  \label{fig: lstm}
\end{figure}

Based on ConvLSTM, we propose a network trained in an encoder-decoder manner for motion prediction and feature extraction. Due to its inherent convolutional capability, this structure is able to tackle the spatio-temporal sequence forecasting problem \cite{xingjian2015convolutional}. The details of this architecture are shown in Fig. \ref{fig: lstm}. $h_{i,j}$ is the hidden state and $c_{i,j}$ is the cell state. The subscript $i$ denotes the time step and $j$ indicates the layer number. All the states with $i=0$ are initialized by the network at the beginning.

The input is a clip of five optical flow images $X_i$ ($i<5$, $i \in \mathbb{Z}$). The rationale for choosing five as the input length is to gain an uniformly sampled clip for one second (30 frames) up to five second (150 frames). More specifically, ``uniformly" means that the interval $L$ between each input is equal. The output of the decoder is the predicted frame in the $L$-frame future. The decoder is in fact a point-wise convolutional layer here, which differs our architecture from other previous work \cite{xingjian2015convolutional, srivastava2015unsupervised}. In this way, motion information of the five-frame input, which can be used for  future motion prediction, is compacted by the encoder. The encoder is regarded as the motion feature extractor, thus, the role of the decoder should be weakened.

The convolution information of the network is shown as in Table. \ref{tab:lstm_arch}. In the third column, the size of the output of every layer is shown. The size has four dimensions: the first dimension is the time step; the second one is the channel number, and the last two refer to the height and width of the input image, respectively. Every ConvLSTM cell takes one frame at one time step, so the first dimension changes to one after the input layer. Additionally, it is worth mentioning that the output from the encoder is the feature needed for maneuver anticipation.

\begin{table}[t!]
	\centering
	\caption{The convolution information about the future motion prediction module}
	\label{tab:lstm_arch}
	\begin{tabular}{lcl}
		\specialrule{.15em}{.0em}{.3em}
		\textbf{Layer}   & \textbf{Kernel Size / Stride}  & \textbf{Output size} \\ 
		\specialrule{.15em}{.3em}{.1em}
		Input         &           & 5$\times$3$\times$h$\times$w \\
	    Layer 0 & (3,3)/(1,1)  & 1$\times$128$\times$h$\times$w \\
		Layer 1 & (3,3)/(1,1)  & 1$\times$64$\times$h$\times$w    \\
		Layer 2 & (3,3)/(1,1)  & 1$\times$64$\times$h$\times$w    \\
		Layer 3 & (3,3)/(1,1)  & 1$\times$32$\times$h$\times$w    \\
		\specialrule{.1em}{.1em}{.1em}
		Conv    & (1,1)/(1,1) & 1$\times$3$\times$h$\times$w    \\
		\specialrule{.15em}{.1em}{.0em}\\
	\end{tabular}
\end{table}	

\subsection{Maneuver Anticipation Framework}
\label{sec:framework}
The proposed method makes use of two input sources: inside and outside videos, as shown in Fig. \ref{fig: overview}. 
For the traffic videos, the FlowNet 2.0 first takes original frames to produce optical flow images. Then, the optical flow images are fed into the ConvLSTM encoder described in the last section. The output from the encoder is then the 3D dimension feature (32$\times$112$\times$176), which will be processed by multiple convolutional blocks (Conv-Block) before fusion. At the same time, the other branch, a 3D ResNet-50, deals with the driver videos. The main body is consistent with the original network in \cite{hara2018can}. Additionally, we added a dropout layer after the average pooling layer in the end to prevent overfitting. The feature we extracted is the input of the last FC layer in ResNet-50, which is a 2048-dimension vector. The input of the ResNet-50 is a 16-frame clip. 

\begin{figure}[h]
    \centering
    \includegraphics[width=.25\linewidth]{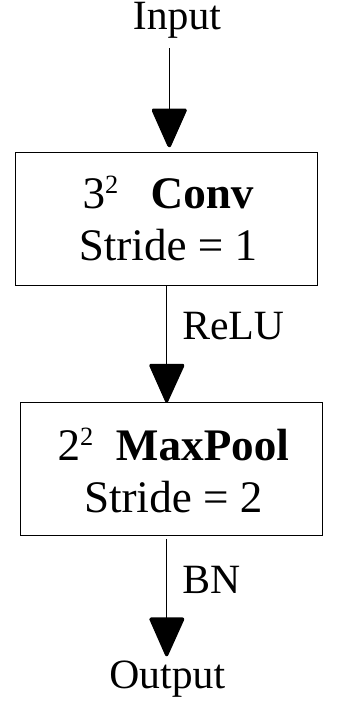}
    \caption{The architecture inside ``Conv-Block"}
    \label{fig:conv-block}
\end{figure}

The novelty of the proposed classifier is that the decoder for outside features is trained jointly with features of inside videos. Its explicit structure is listed in Table. \ref{tab:classifier}. The Conv-Block is for decoding the outside motion. The structure inside one Conv-Block is shown in Fig. \ref{fig:conv-block}, where ``ReLU" refers to the activation function and ``BN" represents the Batch Normalization (BN) layer. There is also a ReLU and a BN between the last two FC layers. The output size after every layer is shown in the third column. In the end, $N_{cls}$ represents the number of classes, which is five in our case.

\begin{table}[t!]
\centering
      \captionof{table}{The architecture of the proposed classifier, which considers joint features from in- and outside videos. The first column indicates the feature source, the second column shows the name of the layer, and the third column is the output size after the layer. The features are combined in the ``Concatenate" layer.}
      \label{tab:classifier}
\begin{tabular}{lcl}
		\specialrule{.15em}{.0em}{.3em}
		\textbf{Feature}   & \textbf{Layer}  & \textbf{Output size} \\ 
		\specialrule{.15em}{.3em}{.1em}
	     & Conv-Block 0 & 64$\times$37$\times$59 \\
	    & Conv-Block 1  & 128$\times$12$\times$20 \\
		Outside & Conv-Block 2  & 256$\times$4$\times$7    \\
		 & Conv-Block 3  & 512$\times$1$\times$2    \\
		\specialrule{.1em}{.1em}{.1em}
		    & Concatenate & 3072$\times$1   \\
		\specialrule{.1em}{.1em}{.1em}
		Both    & FC 0 & 3072$\times$2048   \\
		Both    & FC 1 & 2048$\times$ $N_{cls}$   \\
		Both    & Softmax & $N_{cls}$ \\
		\specialrule{.15em}{.1em}{.0em}\\
\end{tabular}
\end{table}
\section{RESULTS AND DISCUSSIONS}
\label{sec:exp}

\subsection{Dataset}
The Brain4Cars \cite{jain2015car} dataset includes  driver observation videos (1088px $\times$ 1920px, 25 fps) and videos of the outside scenes (480px $\times$ 720px, 30 fps) recorded simultaneously. There are five classes of maneuvers in the dataset: \textit{go straight, left lane change, left turn, right lane change, right turn}.

According to the Brain4cars dataset, the video covers the behavior before the actual maneuver occurs, i.e., no maneuver is performed during the video. In this work, we also study the early detection capability of our models. Therefore, we take every second as a dividing line. In the model evaluation, we give the frames before time step $T$, here $T\in(-5,-4,-3,-2,-1)$. The $-$ represents the time (in second) before the maneuver happens. The shorter videos cover a shorter time period before the maneuver starts. Since the videos have different lengths, we have different amount of input material when we study  early prediction. Moreover, samples with no simultaneous recordings of the inside and outside view are considered as invalid  and not further used in our study. The number of valid video samples for training the whole framework relatively to the covered time period before a maneuver is shown in Table \ref{tab:samples}.  

\begin{table}[h]
\centering
\caption{The number of the valid samples relatively to the video length}
\label{tab:samples}
\begin{tabular}{|l|l|l|l|l|l|}
\hline
video length {[}s{]} & $>4$ & $>3$ & $>2$ & $>1$ & $>0$\\ \hline
samples  &     490    & 542 &   563      &     573    &    585     \\ \hline
\end{tabular}
\end{table}

We use a 5-fold cross-validation for all the experiments in this work, which also aligns with other previous works using the Brain4cars dataset \cite{jain2015car, jainbrain4cars, zhou2018driving, gebert2019end, tonutti2019robust}.
\\

\subsection{Out-cabin Motion Extraction}
For the outside motion feature extraction, we trained the encoder/decoder module presented in \ref{sec:frame prediction}. To achieve a generalized solution, we added a temporal augmentation in training: a 5-frame clip is randomly and uniformly cut and given as the input to the network. The target is the $L$-th frame after the last one in the clip. In the spatial domain, they are first resized to a smaller size (112$\times$176), yet keeping the original scale. Additionally, we employ the Mean Square Error (MSE) as the loss function and Stochastic Gradient Descent (SGD) as the optimizer. The weight decay is set to 0.001 and momentum to 0.9. The whole training takes 60 epochs with the learning rate of 0.1.

\begin{figure}[!htbp]
\centering
\resizebox{.75\linewidth}{!}{
\begin{tikzpicture}
\begin{axis}[xlabel = Interval (frame),
             ylabel = MSE ($\cdot 10^{-3}$ ), 
             ytick={10,12,14,16,18,20},
             yticklabels={10,12,14,16,18,20},
             xtick={5,10,15,20,25,30},
             xticklabels={5,10,15,20,25,30}]
\addplot[mark=*] coordinates {
(5,  10.94)
(10,  14.59)
(15,  16.82)
(20,  18.32)
(25,  18.39)
(30,  18.39)
};
\end{axis}
\end{tikzpicture}
}
\caption{MSE for different interval values}
\label{fig:MSE}
\end{figure}
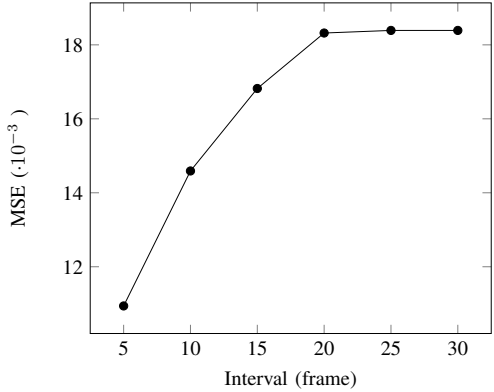

For evaluation, we first studied how far into the future the model is able to predict. More specifically, we evaluated our model with respect to the interval of $L \in (5,10,15,20,25,30)$ frames. As the output of the decoder is the predicted motion in the $L$-th frame after the last input, a larger interval represents a further future. The maximal interval value is $30$ (requiring thus 150 frames), which reaches the maximal video length (5s) in the dataset. On the other hand, an interval less than 5 frames (0.33s) is too short, and thus not considered here. The target frame is the last frame in the video, whereas the metric for comparison is the MSE. The average MSE with respect to different intervals is shown in the Fig. \ref{fig:MSE}. 

Please note that the MSE value is multiplied by 1000 to make the differences more clear. Our results show that it is difficult for the model to predict a far future frame: The model does not learn properly when the interval is larger than 20 frames (0.67s). In order to have relatively precise motion features, we choose the model with $L$ of 5.
After setting the interval $L$ to 5, we evaluated our model with  regard to different time periods of the video. More specifically, the input frames are all included in the time period before $T$ ($T\in(-4,-3,-2,-1,0)$), and the last frame of every second is the target frame. To quantify the comparison between the target and predicted image, we employed three metrics: MSE, Structural Similarity (SSIM) index, and Peak Signal-to-Noise Ratio (PSNR). The results of prediction are shown in Table \ref{tab:result_lstm}. For the PSNR and SSIM, higher values are better. The results of five folds are shown in the form: ``Average (Avg) $\pm$ Standard Error (SE)".

Our results show that the best maneuver prediction is achieved from video information 4 to 5 seconds before the actual maneuver occurs. Thus, motion changes are not massive earlier on before $-3$ second. In case of large motion changes (e.g., when the car is turning), it is hard for the encoder to catch the whole change. Accordingly, in the third and the last second before a maneuver, the outside motion changes noticeably. However, from $-2s$ to $-1s$, motion keeps changing but not as distinct as its contiguous time steps. In general, the important traffic motion changes can be observed within three seconds before the maneuver, which also corresponds to the early detection results in the Section \ref{sec:feature fusion}, where the encoder was emplyed to extract the outside motion features.

\begin{table}[h]
\centering
\caption{Results of future motion prediction.}
\label{tab:result_lstm}
\renewcommand{\arraystretch}{2}
\begin{tabular}{|c|c|c|c|}
\hline
\textbf{prediction at {[}s{]}} &\textbf{MSE ($\cdot$$10^{-3}$)} & \textbf{SSIM} & \textbf{PSNR} \\ \hline
 -4    & $9.13 \pm 0.42$  &  $0.909 \pm 0.001$  & $21.77 \pm 0.16$ \\ \hline
 -3 &  $9.42 \pm 0.40  $  &  $0.906 \pm 0.002$  & $21.49 \pm 0.10$ \\ \hline
-2 &  $10.75 \pm 0.61 $   &  $0.904 \pm 0.002$ & $21.35 \pm 0.18$  \\ \hline
-1 & $ 9.97 \pm 0.22 $  &$0.900 \pm 0.001$ & $21.27 \pm 0.05$\\ \hline
0 & $10.73 \pm 0.46$  & $0.898 \pm 0.002$ & $21.08 \pm 0.10$\\\hline
\end{tabular}
\end{table}

\begin{figure}[h]
\begin{subfigure}{.5\textwidth}
  \centering
  \includegraphics[width=.7\linewidth,frame]{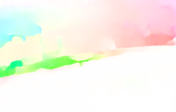}  
  \caption{Target image}
  \label{fig:target}
\end{subfigure}
\begin{subfigure}{.5\textwidth}
  \centering
  \includegraphics[width=.7\linewidth, frame]{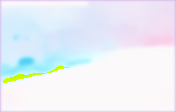}  
  \caption{Predicted image}
  \label{fig:output}
\end{subfigure}
\caption{The comparison of target and the predicted image}
\label{fig:predicted}
\end{figure}

Fig. \ref{fig:output} shows an example of the predicted frame using the proposed encoder/decoder module compared to the target image in \ref{fig:target}. From the visual image results, it is apparent that the major problem is the color disorder. The area in light yellow and the green color is mistaken by light blue in the output. According to the optical flow color coding \cite{baker2011database}, the direction changes 90 degree (from bottom side to right side) from the light yellow to blue, and the green is in between. This detailed motion is difficult for the encoder to catch. 

Using the features extracted from the outside videos by the ConvLSTM-encoder alone can also produce a prediction among five classes. The results are presented in Table \ref{tab:results}, whereas a comparison to related approaches is provided in Table \ref{tab:result_comparison}.

\subsection{In-cabin Action Recognition}
We employ the 3D ResNet-50 for the inside feature extraction, since the 3D ResNet has shown high performance in human action recognition tasks \cite{hara2018can}. However, end-to-end training requires a large amount of the dataset, which is not the case for Brain4cars. Hence, we use the Kinetics-pretrained 3D ResNet-50 \cite{hara2018can} and fine-tune the model with Brain4cars inside videos. 

To prevent overfitting, we added spatial and temporal data augmentation. With regard to spatial augmentation, we added a random crop (but with the focus on the driver side), a random scale and a horizontal flip. It is worth noticing that the label also needs to change accordingly when it is related to the direction (left/right). For temporal augmentation, we randomly but uniformly cut a short clip from every second. The short clips constitutes a 16-frame clip as the input to the 3D ResNet-50, and the input size is 112$\times$112. One extra dropout layer is added before the last FC layer when training. We use a dropout rate of 0.5 an cross entropy loss as out loss function. The model is trained for 60 epochs, with learning rate starting with 0.1 and a decay rate of 0.1 after the 30th and 50th epoch. The optimizer is the SGD with the momentum and weight decay of 0.9 and 0.001, respectively. In out evaluation, we use the frames from the end of every second before $T$ ($T\in(-4,-3,-2,-1,0)$) to compose the 16-frame input for the 3D ResNet.

The main body of trained 3D ResNet-50 is used as the feature extractor. The feature before the last FC layer is fed into the final classifier. The results of using only this module (inside video) for classification are shown in Table \ref{tab:results}, whereas the comparison to related approaches is given in Table \ref{tab:result_comparison}.

\subsection{Feature Fusion}
\label{sec:feature fusion}
After training the ConvLSTM model and 3D ResNet-50 model separately, the features from inside and outside video are extracted by the two trained modules. The obtained outside feature is a volume with the shape of $32\times112\times176$, and the inside feature is a $2048$-size vector. They are fed into the classifier introduced in the section \ref{sec:framework}. We conducted the evaluation procedure with regard to different time periods as in both modules.

The performance indicators are accuracy and the $F_1$-score. The $F_1$-score takes both precision ($Pr$) and recall ($Re$)  of a classifier into consideration (Eq. \ref{f1 eq}). $n$ refers to the number of classes, and $\Omega$ is the set of all the classes that our model can recognize, which includes four maneuvers plus ``no maneuver" class. $TP_i$ indicates the amount of correctly recognized samples of class $i$. $P_i$ and $N_i$ are the number of samples that are predicted as class $i$ and that are labeled as class $i$, separately. 
\begin{equation}
\label{f1 eq}
\begin{aligned}
\centering
&Pr = \frac{1}{n}\sum_{i\in\Omega}{\frac{TP_i}{P_i}}\\
&Re = \frac{1}{n}\sum_{i\in\Omega}{\frac{TP_i}{N_i}}
\\
&F_1 = \frac{2\cdot Pr\cdot Re}{Pr+Re}\\
\end{aligned}
\end{equation}

Table \ref{tab:results} shows the results of accuracy and $F_1$ in \% for different times before the occurrence of a maneuver using different data sources. Both accuracy and $F_1$ increase as the time approaches the beginning of maneuver, despite of different data sources. Intuitively, the early stage of all the maneuvers (or no maneuver) is similar, which is ``going straight". In this case, the longer period the model observes, the more accurate the decision it can make. According to these results, early detection is possible. For example, 71.72\% of the maneuvers are correctly predicted two seconds before the maneuver happens when using both video streams.
\begin{table}[!htbp]
	\centering
    \caption{The results of using proposed framework with different input data sources. The results of five folds are shown in the form: ``Avg $\pm$ SE".}
    \label{tab:results}
    \renewcommand{\arraystretch}{1.5}
	\begin{tabular}{lccl}
		\specialrule{.15em}{.3em}{.3em}
		\textbf{Inside video} & Time period & Acc (\%)  & $F_1$ (\%) \\
		\specialrule{.15em}{.1em}{.0em}
		& {[}-5,-4{]} & 56.49 $\pm$ 0.02 & 48.19 $\pm$ 0.03 \\
		& {[}-5,-3{]} & 63.63 $\pm$ 0.02 & 58.46 $\pm$ 0.02 \\
		 & {[}-5,-2{]} & 70.48 $\pm$ 0.02 & 68.63 $\pm$ 0.03  \\
		& {[}-5,-1{]} & 75.73 $\pm$ 0.01 & 73.09 $\pm$ 0.01  \\
		 &{[}-5,0{]}& 77.40 $\pm$ 0.02 & 75.49 $\pm$ 0.02  \\
		\specialrule{.15em}{.1em}{.0em}

		\textbf{Outside video} & Time period & Acc (\%)  & $F_1$ (\%) \\
		\specialrule{.15em}{.1em}{.0em}
		& {[}-5,-4{]} & 44.08 $\pm$ 0.01 & 38.91 $\pm$ 0.03 \\
		& {[}-5,-3{]} & 44.22 $\pm$ 0.01 & 38.75 $\pm$ 0.01 \\
		 & {[}-5,-2{]} & 50.43 $\pm$ 0.01 & 46.98 $\pm$ 0.01  \\
		& {[}-5,-1{]} & 59.53 $\pm$ 0.01 & 62.37 $\pm$ 0.01  \\
		 &{[}-5,0{]}& 60.87 $\pm$ 0.01 &  66.38 $\pm$ 0.03  \\
		\specialrule{.15em}{.1em}{.0em}

		\textbf{In- \& outside} & Time period & Acc (\%)  & $F_1$ (\%) \\
		\specialrule{.15em}{.1em}{.0em}
		& {[}-5,-4{]} & 59.13 $\pm$ 0.02 & 53.35 $\pm$ 0.02 \\
		& {[}-5,-3{]} & 64.93 $\pm$ 0.02 & 60.33 $\pm$ 0.01 \\
		 & {[}-5,-2{]} & 72.07 $\pm$ 0.02 & 70.56 $\pm$ 0.02  \\
		& {[}-5,-1{]} & 79.92 $\pm$ 0.02 & 78.90 $\pm$ 0.01  \\
		 &{[}-5,0{]}& 83.98 $\pm$ 0.01 & 84.30 $\pm$ 0.01  \\
		\specialrule{.15em}{.3em}{.3em}
	\end{tabular}
\end{table}

\begin{figure*}[]
\begin{subfigure}{.3\textwidth}
  \centering
  \includegraphics[width=.9\linewidth,frame]{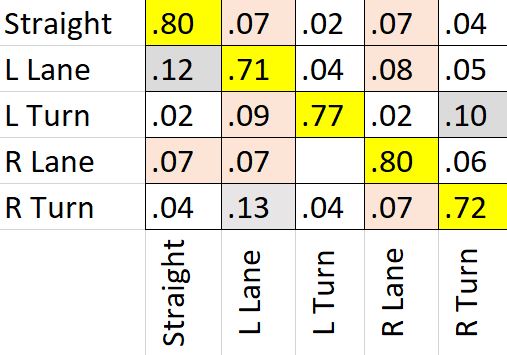}  
  \caption{Inside videos}
  \label{fig:in}
\end{subfigure}
\begin{subfigure}{.3\textwidth}
  \centering
  \includegraphics[width=.9\linewidth, frame]{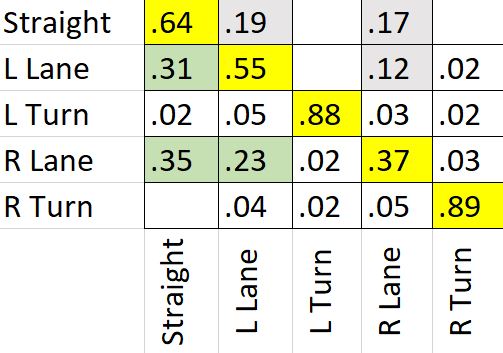}  
  \caption{Outside videos}
  \label{fig:out}
\end{subfigure}
\begin{subfigure}{.3\textwidth}
  \centering
  \includegraphics[width=.9\linewidth, frame]{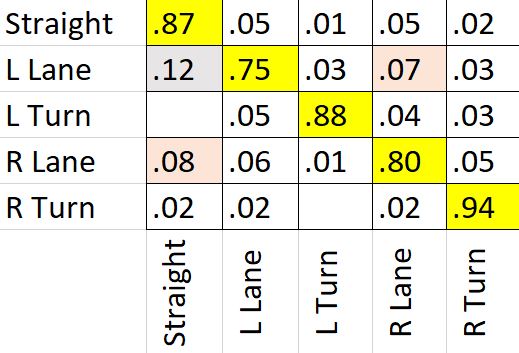}  
  \caption{In and outside videos}
  \label{fig:inout}
\end{subfigure}
\caption{The confusion matrix of using different video streams. The prediction is made at the last second before the occurrence of a maneuvers.}
\label{fig:confusion matrix}
\end{figure*}

The best results are achieved by using both video sources in all different time periods. Only using outside videos gives the worst results when compared to other two data sources. The reason for the poor performance of outside data is that the auto-encoder only provides the motion feature of one future frame. However, the inside feature contains the information over a long time period. Moreover, we can see the decisive motion occurs ordinarily within three seconds before maneuvers. Especially from $-4$ to $-2$, the improvement of accuracy and $F_1$ are substantial. 

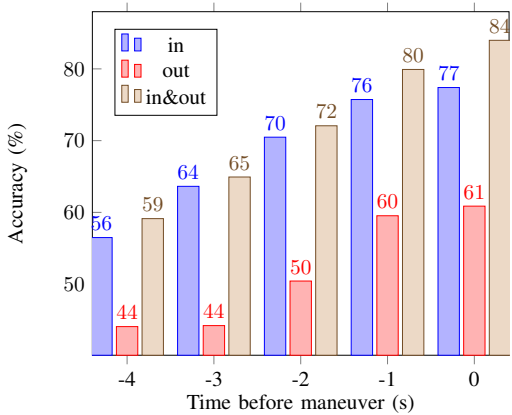
\begin{figure}[!htbp]
        \begin{tikzpicture}[scale=.8]
\pgfplotsset{
    discard if not/.style 2 args={
        x filter/.code={
            \edef\tempa{\thisrow{#1}}
            \edef\tempb{#2}
            \ifx\tempa\tempb
            \else
                \def\pgfmathresult{inf}
            \fi
        }
    }
}

            \begin{axis}[
            ybar,
            symbolic x coords={-4,-3,-2,-1,0},
            xtick=data,
            nodes near coords={
        \pgfmathprintnumber[precision=0]{\pgfplotspointmeta}
       },
            nodes near coords align={vertical},
            legend style={
        at={(0.3,0.95)}},
        xlabel = Time before maneuver (s),
        ylabel = Accuracy (\%),
        ]
        \addplot table [discard if not={group}{in}, x=x, y=y, col sep=comma] {data.csv};
        \addplot table [discard if not={group}{out}, x=x, y=y, col sep=comma] {data.csv};
        \addplot table [discard if not={group}{in&out}, x=x, y=y, col sep=comma] {data.csv};
        \legend{in,out,in\&out}
        \end{axis}
        \end{tikzpicture}
\caption{Accuracy: comparison using different data sources.}
\label{fig:acc}
\end{figure}

\begin{figure}[!htbp]
        \begin{tikzpicture}[scale=.8]
\pgfplotsset{
    discard if not/.style 2 args={
        x filter/.code={
            \edef\tempa{\thisrow{#1}}
            \edef\tempb{#2}
            \ifx\tempa\tempb
            \else
                \def\pgfmathresult{inf}
            \fi
        }
    }
}

            \begin{axis}[
            ybar,
            symbolic x coords={-4,-3,-2,-1,0},
            xtick=data,
            nodes near coords={
        \pgfmathprintnumber[precision=0]{\pgfplotspointmeta}
       },
            nodes near coords align={vertical},
            legend style={
        at={(0.3,0.95)}},
        xlabel = Time before maneuver (s),
        ylabel = $F_1$-score (\%),
        ]
        \addplot table [discard if not={group}{in}, x=x, y=y, col sep=comma] {data2.csv};
        \addplot table [discard if not={group}{out}, x=x, y=y, col sep=comma] {data2.csv};
        \addplot table [discard if not={group}{in&out}, x=x, y=y, col sep=comma] {data2.csv};
        \legend{in,out,in\&out}
        \end{axis}
        \end{tikzpicture}
\caption{$F_1$-score: comparison using different data sources.}
\label{fig:f1}
\end{figure}
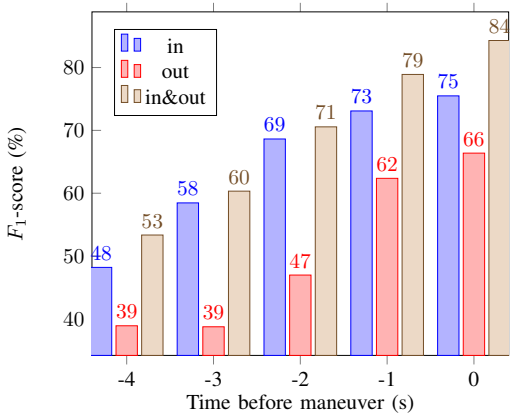

The inside videos always provide good results, but it is still slightly inferior to the joint two-stream input. It is important to see that outside video feature does not depress the performance of the inside video feature, but improves it. Therefore, the information from both inside and outside videos are complementary. Besides, as the outside video become more informative, its effect is more apparent. The differences of accuracy and $F_1$ between inside only and both sides increase steadily after $-3$ seconds. Fig. \ref{fig:acc} and Fig. \ref{fig:f1} illustrate the differences among using different data sources in relation to various time periods more clearly. Additionally, Fig. \ref{fig:confusion matrix} shows the confusion matrix of three models using different data sources. Prediction is made based on time period [-5,0]. From this, an improvement of all classes can be observed when using two video streams.

We compare our results with the ones from work \cite{gebert2019end} in Table. \ref{tab:result_comparison}, since we all use the end-to-end training and investigate the performance with three different data sources. We compare the accuracy, $F_1$ and the number of parameters of our models. The results listed here are all from zero time-to-maneuver and in 5-fold cross-validation.

Our model surpasses the model in \cite{gebert2019end} except using only inside view. It is because the 3D ResNet-101 is used in \cite{gebert2019end}, which has almost two times more parameters than 3D ResNet-50 in our work. We choose to use a smaller ResNet in order to avoid outfitting problems when fine tuning a very large network with a small dataset. Moreover, a low resource-cost model is preferable for automobile applications. Our framework outperforms the previous work with much less parameters in using two-stream input: It achieves 83.98\% of accuracy and 84.30\% of $F_1$ averagely within five folds, surpassing the previous work by 8.48 percentage points in accuracy and 11.1 percentage points in $F_1$. When only considering outside videos, our models surpasses theirs by 7.67 percentage points and 22.98 percentage points in accuracy and $F_1$, respectively. It achieves to extract useful features from outside with much less parameters. More importantly, our model does not confront the same problem that the outside videos weaken the classifier performance. In other words, our results show that the information from outside videos are also valuable.

\begin{table}[!htbp]
	\centering
	\caption{Comparison of our proposed framework with other method. The results of five folds are shown in the form: ``Avg $\pm$ SE". In order to show a clear difference, we use ``$m$" to represent the number of parameters in FlowNet2.0, which is a common module in both methods.}
	\label{tab:result_comparison}
	\renewcommand{\arraystretch}{1.5}
	\begin{tabular}{l|ccll}
	\specialrule{.15em}{.0em}{.3em}
		Method  & Data Source & Acc (\%) & $F_1$ (\%)& Param.(M)\\
		\specialrule{.15em}{.3em}{.1em}
& inside only & 83.1 $\pm$ 2.5 &  81.7 $\pm$ 2.6 & 85.26+$m$  \\
\cite{gebert2019end} & outside only & 53.2 $\pm$ 0.5 & 43.4 $\pm$ 0.9 & 85.26+$m$ \\
                    & in-\&out-side & 75.5 $\pm$ 2.4 & 73.2 $\pm$ 2.2 & 170.52+$m$ \\
		\specialrule{.15em}{.3em}{.1em}
		    & inside only & 77.40 $\pm$ 0.02 & 75.49 $\pm$ 0.02 & 46.22\\
		our & outside only & 60.87 $\pm$ 0.01\ & 66.38 $\pm$ 0.03 &5.41+$m$\\
	       & in-\&outside & \textbf{83.98} $\pm$ \textbf{0.01} & \textbf{84.30} $\pm$ \textbf{0.01} & \textbf{57.92+$m$}\\
		\specialrule{.15em}{.1em}{.0em}
	\end{tabular}
\end{table}	

\begin{figure}[h]
\centering
\begin{tikzpicture}[scale=.7]
\pgfplotsset{
    discard if not/.style 2 args={
        x filter/.code={
            \edef\tempa{\thisrow{#1}}
            \edef\tempb{#2}
            \ifx\tempa\tempb
            \else
                \def\pgfmathresult{inf}
            \fi
        }
    }
}

        \begin{axis}[
                    legend style={
        at={(1.19,0.41)}},
                    ylabel = $F_1$-score (\%),
                    xlabel = Threshold,]
        \addplot table [discard if not={group}{5}, x=x, y=y, col sep=comma] {data3.csv};
        \addplot table [discard if not={group}{4}, x=x, y=y, col sep=comma] {data3.csv};
        \addplot table [discard if not={group}{3}, x=x, y=y, col sep=comma] {data3.csv};
        \addplot table [discard if not={group}{2}, x=x, y=y, col sep=comma] {data3.csv};
        \addplot table [discard if not={group}{1}, x=x, y=y, col sep=comma] {data3.csv};
        \legend{5s, 4s, 3s, 2s, 1s}
        \end{axis}
        \end{tikzpicture}
\caption{Effect of using thresholds. Two-stream input with different video lengths (from 1 to 5 seconds).}
\label{fig:threshold}
\end{figure}
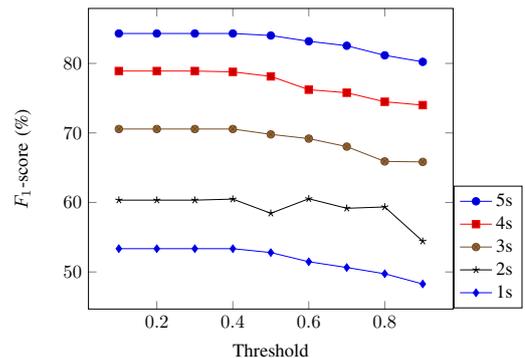

We also conduct an experiment using similar threshold policy as in \cite{jain2015car, jainbrain4cars} on our model which uses two-stream video: If the probability is NOT greater than the threshold, then \textit{``go straight"} is predicted. As shown in Fig. \ref{fig:threshold}, the performance gets worse when this threshold is larger than $0.4$ in all lengths of input videos, since the model is trained on a balanced loss function and learns motion features of all five maneuvers. It always gives a relatively confident prediction with a probability over $0.4$. For our model, no threshold policy is necessary. 

\section{CONCLUSION AND FUTURE WORK}
In this work, we propose a framework that considers both inside and outside cabin motion features to anticipate the driver maneuver intention. We propose to extract the outside traffic motion using a ConvLSTM-based auto-encoder. These motion features are decoded by a novel classifier architecture, which considers the in- and outside motions jointly. Our model is trained in end-to-end style, without using any manual-encoded or hand-crafted features. 
Our results show that dual input (driver observation and driving scene videos) surpasses by far related approaches based on single input analyses. Additionally, we validate  experimentally that both inside and outside videos convey valuable and complementary information. This conclusion suggests that both traffic scenes and driver behaviors should be taken into consideration when anticipating maneuver intention.
                           
For our future work, we plan to improve the performance of the outside motion decoder in the classifier by training a more delicate decoder which can interpret the motion covering a longer time period. In this way, the module would gain a perspective of the entire outside motion. Moreover, accurately predicting the motion of the further future is another aim for our future work. 

\section*{ACKNOWLEDGMENT}
Funded by the Deutsche Forschungsgemeinschaft (DFG, German Research Foundation) under Germany’s Excellence Strategy – EXC-Number 2064/1 – Project number 390727645.









\begin{thebibliography}{}

\bibitem{jain2015car}
Jain, Ashesh and Koppula, Hema S and Raghavan, Bharad and Soh, Shane and Saxena, Ashutosh, \emph{Car that knows before you do: Anticipating maneuvers via learning temporal driving models}, Proceedings of the IEEE International Conference on Computer Vision, pages 3182--3190, 2015

\bibitem{WHO}
\emph{Road traffic injuries},
February 7. 2020. Accessed on: Feb. 9, 2020. [Online]. Available: https://www.who.int/news-room/fact-sheets/detail/road-traffic-injuries

\bibitem{morando2018studying}
Morando, Mark Mario and Tian, Qingyun and Truong, Long T and Vu, Hai L, \emph{Studying the safety impact of autonomous vehicles using simulation-based surrogate safety measures}, Journal of advanced transportation, vol.2018, 2018, Hindawi

\bibitem{fox2014self}
Fox, M, \emph{Self-driving cars safer than those driven by humans: Bob Lutz}, CNBC, [Online]. Available: www. cnbc. com, 2014

\bibitem{teoh2017rage}
Teoh, Eric R and Kidd, David G, \emph{Rage against the machine? Google's self-driving cars versus human drivers}, Journal of safety research, vol.63, page 57--60, 2017, Elsevier

\bibitem{dftuk}
\emph{Most Common Causes for Road Accidents in Britain Revealed}, July 2. 2018. Accessed on: Feb.9,2020. [Online]. Available: https://www.regtransfers.co.uk/content/common-causes-for-road-accidents-in-britain/

\bibitem{ramanishka2018toward}
Ramanishka, Vasili and Chen, Yi-Ting and Misu, Teruhisa and Saenko, Kate, \emph{Toward driving scene understanding: A dataset for learning driver behavior and causal reasoning}, Proceedings of the IEEE Conference on Computer Vision and Pattern Recognition, page 7699--7707, 2018

\bibitem{xingjian2015convolutional}
Xingjian, SHI and Chen, Zhourong and Wang, Hao and Yeung, Dit-Yan and Wong, Wai-Kin and Woo, Wang-chun, \emph{Convolutional LSTM network: A machine learning approach for precipitation nowcasting}, Advances in neural information processing systems, page 802--810, 2015


\bibitem{jainbrain4cars}
Jain, Ashesh and Soh, Shane and Raghavan, Bharad and Singh, Avi and Koppula, Hema S and Saxena, Ashutosh, \emph{Brain4Cars: Sensory-Fusion Recurrent Neural Models for Driver Activity Anticipation}

\bibitem{zhou2018driving}
Zhou, Dong and Ma, Huimin and Dong, Yuhan, \emph{Driving maneuvers prediction based on cognition-driven and data-driven method}, 2018 IEEE Visual Communications and Image Processing (VCIP), page 1--4, 2018, IEEE

\bibitem{gebert2019end}
Gebert, Patrick and Roitberg, Alina and Haurilet, Monica and Stiefelhagen, Rainer, \emph{End-to-end Prediction of Driver Intention using 3D Convolutional Neural Networks}, 2019 IEEE Intelligent Vehicles Symposium (IV), page 969--974, 2019, IEEE

\bibitem{tonutti2019robust}
Tonutti, Michele and Ruffaldi, Emanuele and Cattaneo, Alessandro and Avizzano, Carlo Alberto, \emph{Robust and subject-independent driving manoeuvre anticipation through Domain-Adversarial Recurrent Neural Networks}, Robotics and Autonomous Systems, vol.115, page 162--173, 2019, Elsevier

\bibitem{menze2015object}
Menze, Moritz and Geiger, Andreas, \emph{Object scene flow for autonomous vehicles}, Proceedings of the IEEE conference on computer vision and pattern recognition, page 3061--3070, 2015

\bibitem{carreira2017quo}
Carreira, Joao and Zisserman, Andrew, \emph{Quo vadis, action recognition? a new model and the kinetics dataset}, proceedings of the IEEE Conference on Computer Vision and Pattern Recognition, page 6299--6308, 2017

\bibitem{hara2018can}
Hara, Kensho and Kataoka, Hirokatsu and Satoh, Yutaka, \emph{Can spatiotemporal 3d cnns retrace the history of 2d cnns and imagenet?}, Proceedings of the IEEE conference on Computer Vision and Pattern Recognition, page 6546--6555, 2018

\bibitem{horn1981determining}
Horn, Berthold KP and Schunck, Brian G, \emph{Determining optical flow}, Techniques and Applications of Image Understanding, vol.281, page 319--331, 1981, International Society for Optics and Photonics

\bibitem{adelson1985spatiotemporal}
Adelson, Edward H and Bergen, James R, \emph{Spatiotemporal energy models for the perception of motion}, Josa a, vol.2, page 284--299, 1985, Optical Society of America

\bibitem{anandan1989computational}
Anandan, Padmanabhan, \emph{A computational framework and an algorithm for the measurement of visual motion}, International Journal of Computer Vision, vol.2, page 283--310, 1989, Springer

\bibitem{dosovitskiy2015flownet}
Dosovitskiy, Alexey and Fischer, Philipp and Ilg, Eddy and Hausser, Philip and Hazirbas, Caner and Golkov, Vladimir and Van Der Smagt, Patrick and Cremers, Daniel and Brox, Thomas, \emph{Flownet: Learning optical flow with convolutional networks}, Proceedings of the IEEE international conference on computer vision, page 2758--2766, 2015

\bibitem{ilg2017flownet}
Ilg, Eddy and Mayer, Nikolaus and Saikia, Tonmoy and Keuper, Margret and Dosovitskiy, Alexey and Brox, Thomas, \emph{Flownet 2.0: Evolution of optical flow estimation with deep networks}, Proceedings of the IEEE conference on computer vision and pattern recognition, page 2462--2470, 2017

\bibitem{baker2011database}
Baker, Simon and Scharstein, Daniel and Lewis, JP and Roth, Stefan and Black, Michael J and Szeliski, Richard, \emph{A database and evaluation methodology for optical flow}, International journal of computer vision, vol.92, page 1--31, 2011, Springer

\bibitem{zhou2016learning}
Zhou, Bolei and Khosla, Aditya and Lapedriza, Agata and Oliva, Aude and Torralba, Antonio, \emph{Learning deep features for discriminative localization}, Proceedings of the IEEE conference on computer vision and pattern recognition, page 2921--2929, 2016

\bibitem{mikolov2010recurrent}
Mikolov, Tom{\'a}{\v{s}} and Karafi{\'a}t, Martin and Burget, Luk{\'a}{\v{s}} and {\v{C}}ernock{\`y}, Jan and Khudanpur, Sanjeev, \emph{Recurrent neural network based language model},Eleventh annual conference of the international speech communication association, 2010

\bibitem{sutskever2014sequence}
Sutskever, Ilya and Vinyals, Oriol and Le, Quoc V, \emph{Sequence to sequence learning with neural networks}, Advances in neural information processing systems, page 3104--3112, 2014

\bibitem{wu2015modeling}
Wu, Zuxuan and Wang, Xi and Jiang, Yu-Gang and Ye, Hao and Xue, Xiangyang, \emph{Modeling spatial-temporal clues in a hybrid deep learning framework for video classification}, Proceedings of the 23rd ACM international conference on Multimedia, page 461--470, 2015

\bibitem{srivastava2015unsupervised}
Srivastava, Nitish and Mansimov, Elman and Salakhudinov, Ruslan, \emph{Unsupervised learning of video representations using lstms}, International conference on machine learning, page 843--852, 2015

\bibitem{braunagel2015driver}
Braunagel, Christian and Kasneci, Enkelejda and Stolzmann, Wolfgang and Rosenstiel, Wolfgang, \emph{Driver-activity recognition in the context of conditionally autonomous driving}, 2015 IEEE 18th International Conference on Intelligent Transportation Systems, page 1652--1657, 2015

\bibitem{braunagel2017online}
Braunagel, Christian and Geisler, David and Rosenstiel, Wolfgang and Kasneci, Enkelejda, \emph{Online recognition of driver-activity based on visual scanpath classification}, IEEE Intelligent Transportation Systems Magazine, 9 (4), page 23--36, 2017

\bibitem{braunagel2017ready}
Braunagel, Christian and Rosenstiel, Wolfgang and Kasneci, Enkelejda, \emph{Ready for take-over? A new driver assistance system for an automated classification of driver take-over readiness}, IEEE Intelligent Transportation Systems Magazine, 9:(4), page 10--22, 2017

\end{thebibliography}
\end{document}